\documentclass[runningheads]{llncs}
\usepackage{graphicx}
\usepackage{xcolor}
\usepackage{bm}
\usepackage{amsmath}
\usepackage{subcaption}
\usepackage{booktabs,tabularx}
\usepackage{upgreek}
\begin{document}
\title{Inertial Measurements for Motion Compensation in Weight-bearing Cone-beam CT of the Knee}
\titlerunning{Inertial Measurements for CT Motion Compensation}
%
\author{Jennifer~Maier\inst{1,2}\and%
Marlies~Nitschke\inst{2}\and%
Jang-Hwan~Choi\inst{3}\and%
Garry~Gold\inst{4}\and%
Rebecca~Fahrig\inst{5}\and%
Bjoern~M.~Eskofier\inst{2}\and%
Andreas~Maier\inst{1}}
\authorrunning{J. Maier et al.}
%
\institute{Pattern Recognition Lab, Friedrich-Alexander-Univerist\"at Erlangen-N\"urnberg (FAU), Erlangen, Germany \and Machine Learning and Data Analytics Lab, Friedrich-Alexander-Univerist\"at Erlangen-N\"urnberg (FAU), Erlangen, Germany \and College of Engineering, Ewha Womans University, Seoul, Korea \and Stanford University, Stanford, California, USA \and Siemens Healthcare GmbH, Forchheim, Germany \\\email{jennifer.maier@fau.de}}
\maketitle              
\begin{abstract}
Involuntary motion during weight-bearing cone-beam computed tomography (CT) scans of the knee causes artifacts in the reconstructed volumes making them unusable for clinical diagnosis.
Currently, image-based or marker-based methods are applied to correct for this motion, but often require long execution or preparation times.
We propose to attach an inertial measurement unit (IMU) containing an accelerometer and a gyroscope to the leg of the subject in order to measure the motion during the scan and correct for it.
To validate this approach, we present a simulation study using real motion measured with an optical 3D tracking system.
With this motion, an XCAT numerical knee phantom is non-rigidly deformed during a simulated CT scan creating motion corrupted projections.
A biomechanical model is animated with the same tracked motion in order to generate measurements of an IMU placed below the knee.
In our proposed multi-stage algorithm, these signals are transformed to the global coordinate system of the CT scan and applied for motion compensation during reconstruction.
Our proposed approach can effectively reduce motion artifacts in the reconstructed volumes.
Compared to the motion corrupted case, the average structural similarity index and root mean squared error with respect to the no-motion case improved by 13-21\% and 68-70\%, respectively.
These results are qualitatively and quantitatively on par with a state-of-the-art marker-based method we compared our approach to.
The presented study shows the feasibility of this novel approach, and yields promising results towards a purely IMU-based motion compensation in C-arm CT.
\end{abstract}
\section{Introduction}
\begin{figure}[t]
  \begin{subfigure}[b]{0.6\textwidth}
    \centering
    \includegraphics[height=5cm]{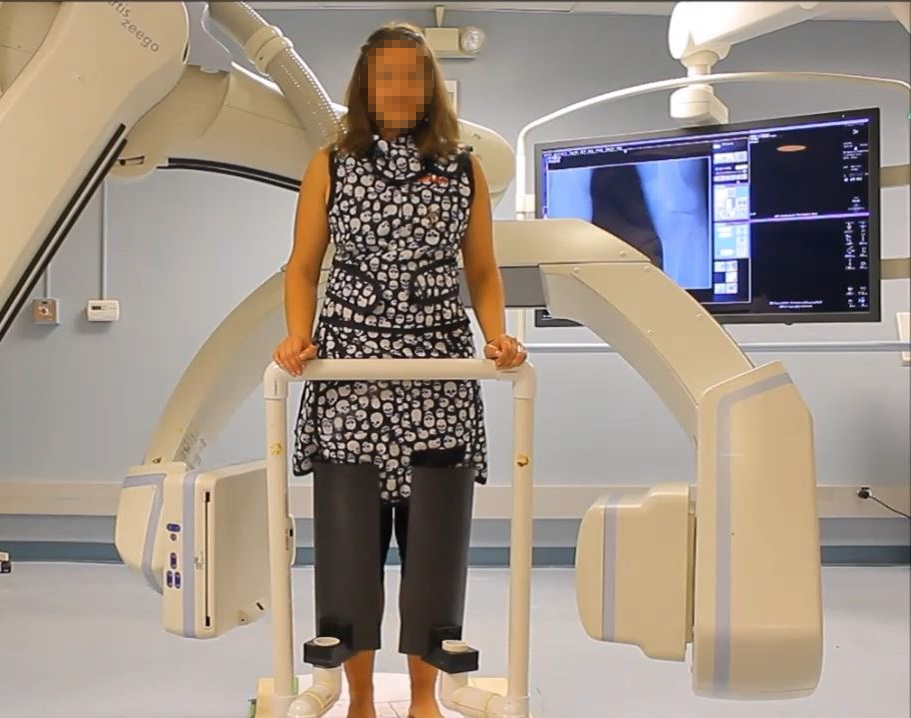}
    \caption{Horizontal C-arm CT scan.}
    \label{fig-CArmScan}
  \end{subfigure}
  \begin{subfigure}[b]{0.35\textwidth}
    \centering
    \includegraphics[height=5cm]{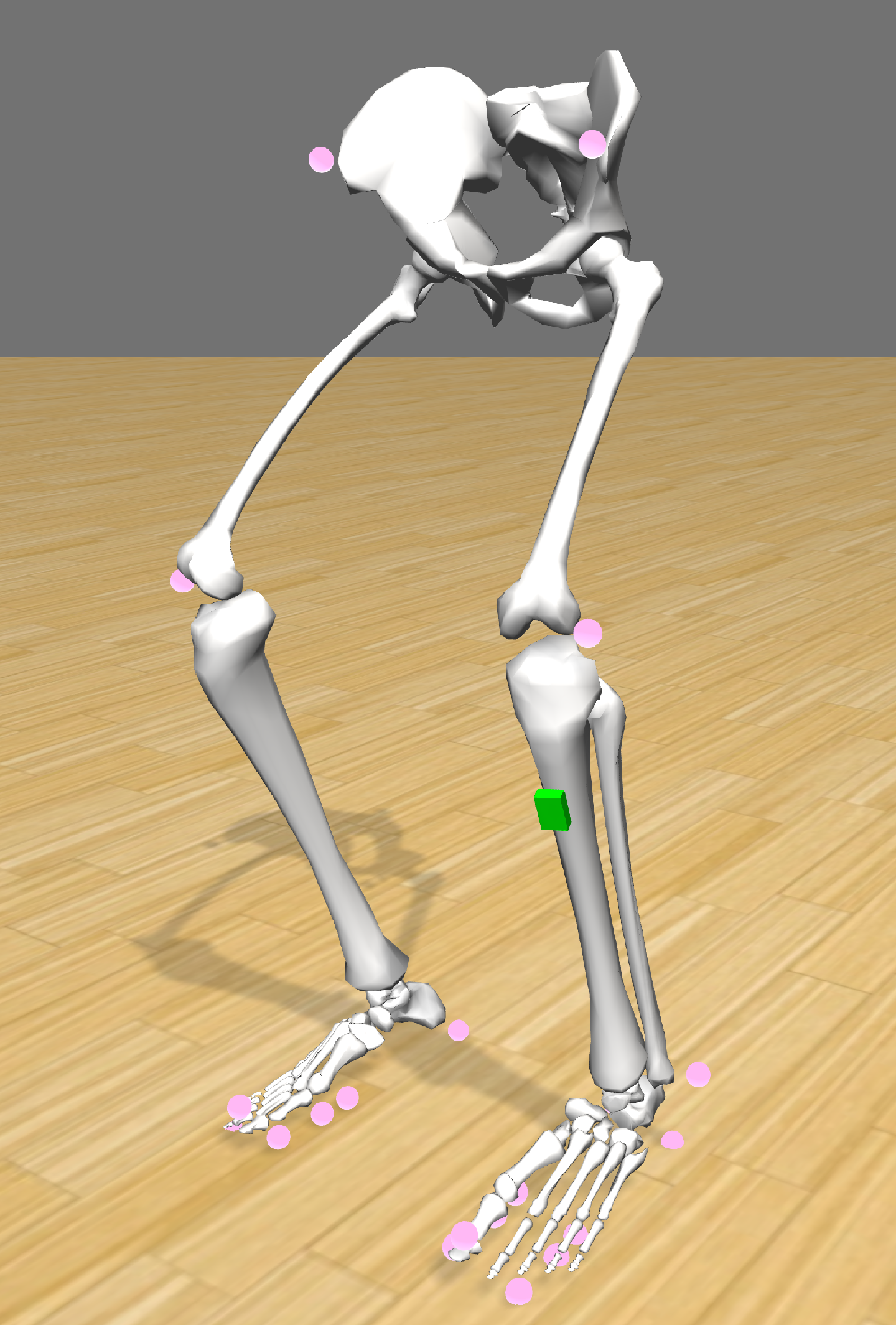}
    \caption{Biomechanical model.}
    \label{fig-OpenSimModel}
  \end{subfigure}
  \caption{(a) Setting of a weight-bearing C-arm CT acquisition during which subjects move involuntarily. (b) Biomechanical model used for motion modeling in OpenSim and for simulation of inertial measurements. The tracked markers are shown in pink and the simulated sensor is shown in green.}
\end{figure}

Osteoarthritis is a disease affecting articular cartilage in the joints, leading to a higher porosity and eventually to loss of tissue \cite{Arden2006}.
The structural change of cartilage also has an influence on its mechanical properties, i.e. its behavior when put under stress \cite{Powers2003}.
To analyze how diseased cartilage in the knee joint changes under stress compared to healthy tissue, it can be imaged under load conditions.
This can be realized by scanning the knee joint in a weight-bearing standing position using a flexible C-arm cone-beam Computed Tomography (CT) system rotating on a horizontal trajectory, as depicted in Figure~\ref{fig-CArmScan} \cite{Maier2011}.
However, standing subjects will show more involuntary motion due to body sway when naturally standing compared to the conventional supine scanning position \cite{Sisniega2016}.
Since standard CT reconstruction assumes stationary objects, this motion leads to streaking artifacts, double contours, and blurring in the reconstructed volumes making them unsuitable for clinical diagnosis.

Multiple approaches to correct for this motion have been proposed in literature.
There exist purely image-based methods, like 2D/3D registration \cite{Berger2016} or the use of a penalized image sharpness criterion \cite{Sisniega2017}, that show very good motion compensation results but are computationally expensive.
It is also possible to use epipolar consistency conditions for motion compensation, but for knee imaging this has so far only been applied for estimating translation and not rotation \cite{Bier2017}.
Recently, also deep learning methods were applied in CT motion correction.
Bier et al. proposed a neural network to detect anatomical landmarks in projection images, but the approach was not robust when other objects were present and was only evaluated for tracking motion \cite{Bier2018landmark}. Its feasibility for compensating motion was not investigated.
An approach requiring external hardware is using range cameras to track motion during the scan, which so far worked well on purely simulated data \cite{Bier2018range}.
The gold standard method for knee motion compensation is based on small metallic markers attached to the scanned leg, and was proposed in \cite{Choi2013,Choi2014}.
The markers tracked in the projections can be used to iteratively estimate motion.
However, the placement of the markers is tedious and the metal produces artifacts in the reconstructions.

Inertial measurement units (IMUs) containing an accelerometer and a gyroscope have found use in C-arm CT for navigation \cite{Jost2016} and calibration \cite{Lemammer2019} purposes.
We propose to use these small and lightweight sensors for motion compensation in C-arm CT.
For this purpose, an IMU is attached to the leg of the subject to measure motion during the scan.
To show the feasibility of this approach, we present a simulation study using real 3D human swaying motion recorded with an optical tracking system.
These measurements are used to animate an OpenSim biomechanical model by inverse kinematics computation \cite{Delp2007,Seth2018}.
The model's movement is on the one hand used to deform an XCAT numerical phantom for the generation of motion corrupted CT projections \cite{Segars2010}.
On the other hand, it is used to simulate inertial measurements of a sensor placed on the leg.
The simulated measurements are processed in a multi-stage motion correction pipeline consisting of gravity removal, local velocity computation, global transformation, and projection geometry correction.
%
%
\section{Materials and Methods}
In order to generate realistic X-ray projections and IMU measurements, the XCAT and OpenSim models are animated with real human swaying motion.
This motion is recorded with a Vicon optical motion capture system (Vicon, Oxford, UK) tracking seven reflective markers attached to the subject's body at a sampling rate of 120\,Hz. 
The markers are placed on the sacrum, and on the right and left anterior superior iliac spine, lateral epicondyle of the knee and malleolus lateralis.
Seven healthy subjects are recorded holding a squat at 30 and 60 degrees of knee flexion.
Afterwards, a biomechanical model of the human lower body based on the model presented in \cite{Hamner2010} is scaled to each subject's anthropometry using the software OpenSim, see Fig.~\ref{fig-OpenSimModel} \cite{Delp2007,Seth2018}.
With this model and the measured 3D marker positions, the inverse kinematics are computed in order to find the generalized coordinates (i.e. global position and orientation of the pelvis and the joint angles) that best represent the measured motion.
Before further processing, the generalized coordinates are filtered with a moving average filter with a span of 60 in order to remove system noise from the actual movement.
The 3D positions of the sacrum, and the left and right hip joint center, knee joint center and ankle joint center over time are extracted from the animated model and used for the XCAT CT projection generation (Section \ref{sec-xcat}).
A virtual sensor is placed on the animated model's shank in order to simulate IMU measurements (Section \ref{sec-imu}), which are used for the motion compensated reconstruction (Section \ref{sec-moco}).
\subsection{Generation of motion corrupted projections}
\label{sec-xcat}
The XCAT numerical phantom is a model of the human body with its legs consisting of the bones tibia, fibula, femur and patella including bone marrow and surrounding body soft tissue \cite{Segars2010}.
The shapes of these structures are defined by non-uniform rational B-Splines (NURBS) and by the positions of their control points.
By shifting these control points for each time step of the simulated CT scan based on the hip, knee and ankle joint centers of the OpenSim model, the upper leg and lower leg of the XCAT model are animated individually leading to a non-rigid motion.
With the deformed model for each time step, X-ray projection images of a simulated CT scan are generated.
The detector of size 620$\times$480\,pixels with isotropic pixel resolution of 0.616\,mm rotates on a virtual circular trajectory with a source detector distance of 1198\,mm and a source isocenter distance of 780\,mm.
The angular increment between projections is 0.8\,degrees and in total 248 projections are generated, corresponding to a sampling rate of 31\,Hz.
Forward projections of the deformed model are created as described in \cite{Maier2012}.
Since a healthy human's knees in a natural standing position are too far apart to both fit on the detector, the rotation center of the scan is placed in the center of the left leg.
\subsection{Simulation of inertial measurements}
\label{sec-imu}
An IMU is a small lightweight device that measures its acceleration and angular velocity on three perpendicular axes.
Additionally, the accelerometer always also senses the earth's gravitational field distributed on its three axes depending on the current orientation.
Such a sensor is virtually placed on the shank 14\,cm below the left knee joint aligned with the shank segment.
The simulated acceleration $\mathbf{a}_{i}$ and angular velocity $\bm{\omega}_{i}$ at time point $i$ are computed as follows \cite{Bogert1996,Desapio2017}:
\begin{equation}
    \mathbf{a}_{i} = \mathbf{R}_i^\top(\ddot{\mathbf{r}}_{Seg,i} + \ddot{\mathbf{R}}_i \mathbf{p}_{Sen,i} - \mathbf{g})
\end{equation}
\begin{equation}
    \bm{\omega}_i = (\omega_{x,i}, \omega_{y,i}, \omega_{z,i})^\top
\end{equation}
\begin{equation}
    [\bm{\omega}_i]_{\times} = \mathbf{R}_i^\top\dot{\mathbf{R}}_i = 
    \begin{pmatrix}
        0 & -\omega_{z,i} & \omega_{y,i} \\
        \omega_{z,i} & 0 & -\omega_{x,i} \\
        -\omega_{y,i} & \omega_{x,i} & 0
    \end{pmatrix}
\end{equation}
The 3$\times$3 rotation matrix $\mathbf{R}_i$ describes the orientation of the sensor at time point $i$ in the global coordinate system, $\dot{\mathbf{R}}_i$ and $\ddot{\mathbf{R}}_i$ are its first and second order derivatives with respect to time. The position of the segment the sensor was mounted on in the global coordinate system at time point $i$ is described by $\mathbf{r}_{Seg,i}$, with $\ddot{\mathbf{r}}_{Seg,i}$ being its second order derivative. $\mathbf{p}_{Sen,i}$ is the position of the sensor in the local coordinate system of the segment the sensor was mounted on. All required parameters are obtained by computing the forward kinematics of the biomechanical model. $\mathbf{g}=(0,-9.80665, 0)^\top$ is the global gravity vector.
\subsection{Motion compensated reconstruction}
\label{sec-moco}
The simulated IMU measurements are used to estimate a rigid motion describing the 3D change of orientation and position from each time step to the next in the global coordinate frame.
The sensor's coordinate system $\mathbf{S}_i$ at each time step $i$, i.e. its orientation and position in the global frame is described by the affine matrix
\begin{equation}
    \mathbf{S}_i =
    \begin{pmatrix}
        \begin{array}{c|c}
              \mathbf{\hat{R}}_{i} & \mathbf{\hat{t}}_{i} \\ 
              \hline
              \mathbf{0} & 1
        \end{array}
    \end{pmatrix},
\end{equation}
where $\mathbf{\hat{R}}_{i}$ is a 3$\times$3 rotation matrix, and $\mathbf{\hat{t}}_{i}$ is a 3$\times$1 translation vector. The initial pose $\mathbf{S}_0$ is assumed to be known.

The gyroscope measures the angular velocity, which is the change of orientation over time on the three axes of the sensor's local coordinate system.
Therefore, this measurement can directly be used to rotate the sensor from each time step to the next.
The measured accelerometer signal, however, needs to be processed to obtain the positional change over time.
First, the gravity measured on the sensor's three axes is removed based on its global orientation.
If $\mathbf{G}_i$ is a 3D rotation matrix containing the rotation change measured by the gyroscope at time step $i$, the global orientation of the sensor is described by
\begin{equation}
    \mathbf{\hat{R}}_{i+1} = \mathbf{\hat{R}}_{i} \mathbf{G}_i.
\end{equation}
The global gravity vector $\mathbf{g}$ is transformed to the sensor's local coordinate system at each time step $i$ using
\begin{equation}
    \mathbf{g}_{i} = \mathbf{\hat{R}}_{i}^\top \mathbf{g}.
\end{equation}
To obtain the gravity-free acceleration $\mathbf{\bar{a}}_{i}$, the gravity component then is removed from $\mathbf{a}_{i}$ by adding the local gravity vector $\mathbf{g}_{i}$ at each time step $i$.
To obtain the sensor's local velocity $\mathbf{v}_{i}$, i.e. its position change over time, the integral of the gravity-free acceleration is computed.
The integration must be performed considering the sensor's orientation changes.
\begin{equation}
    \mathbf{v}_{i+1} = \mathbf{G}_i^\top(\mathbf{\bar{a}}_i + \mathbf{v}_{i}).
\end{equation}
In this study, we assume that the sensor's initial velocity $\mathbf{v}_{0}$ is known.

The local rotational change $\bm{\omega}_i$ and positional change $\mathbf{v}_{i}$ for each time step $i$ are linearly resampled to the CT scan's sampling frequency and rewritten to an affine matrix
\begin{equation}
    \bm{\Updelta}_{l,i} =
    \begin{pmatrix}
        \begin{array}{c|c}
              \mathbf{G}_i & \mathbf{v}_i \\ 
              \hline
              \mathbf{0} & 1
        \end{array}
    \end{pmatrix}.
\end{equation}
To obtain the change in the global coordinate system, $\bm{\Updelta}_{l,i}$ is transformed for each time step using its pose $\mathbf{S}_i$:
\begin{equation}
    \bm{\Updelta}_{g,i} = \mathbf{S}_i\bm{\Updelta}_{l,i}\mathbf{S}_i^{-1}
\end{equation}
\begin{equation}
    \mathbf{S}_{i+1} = \bm{\Updelta}_{g,i}\mathbf{S}_i
\end{equation}

For the motion compensated reconstruction, the first projection is used as reference without motion, so the affine matrix containing rotation and translation is the identity matrix $\mathbf{M}_0 = \mathbf{I}$.
For each subsequent time step, the motion matrix is computed as
\begin{equation}
    \mathbf{M}_{i+1} = \mathbf{M}_{i} \bm{\Updelta}_{g,i}.
\end{equation}
These motion matrices are applied to the projection matrices of the system's geometry to correct for motion.
\subsection{Evaluation}
We reconstruct the projections as volumes of size $512^3$ with isotropic spacing of 0.5\,mm using GPU accelerated filtered back-projection in the software framework CONRAD \cite{Maier2013}.

To evaluate the performance of the proposed method, the resulting reconstructions are compared to a reconstruction without motion correction, and to a ground truth reconstruction from projections where the initial pose of the subject was kept static throughout the scan.
Furthermore, we compare to reconstructions from the gold standard marker-based approach.
For this purpose, small highly attenuating circular markers on the skin are simulated and tracked as proposed in \cite{Choi2014}.
All volumes are scaled from 0 to 1 and registered to the reference reconstruction without motion for comparison.
As metrics for image quality evaluation we compute the root mean squared error (RMSE) and the structural similarity (SSIM).
The SSIM index considers differences in luminance, contrast and structure for comparison and ranges from 0 (no similarity) to 1 (identical images) \cite{Wang2004}.
\section{Results}
\begin{figure}[t]
       \centering
\begin{tabular}{cccc}
\includegraphics[width=0.236\textwidth]{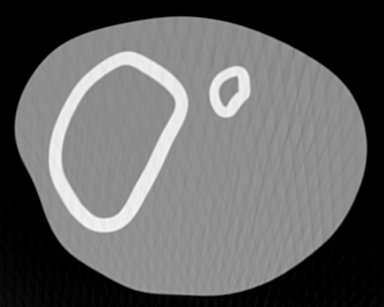}&
\includegraphics[width=0.236\textwidth]{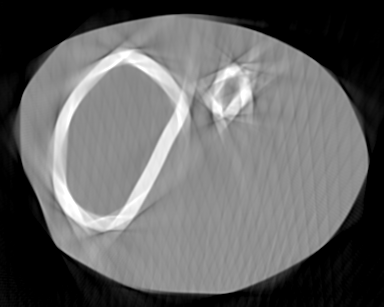}&
\includegraphics[width=0.236\textwidth]{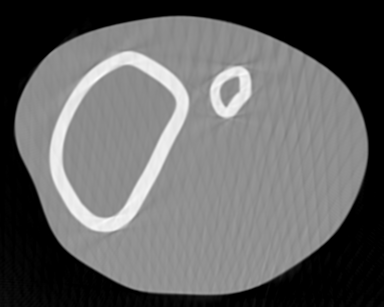}&
\includegraphics[width=0.236\textwidth]{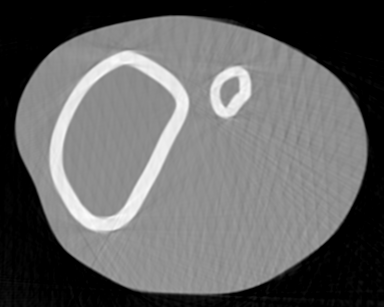}\\

\includegraphics[width=0.236\textwidth]{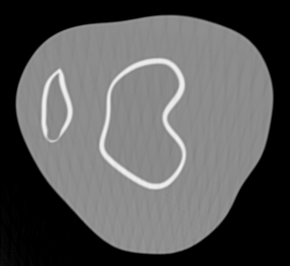}&
\includegraphics[width=0.236\textwidth]{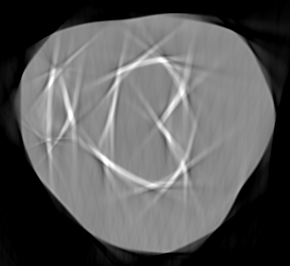}&
\includegraphics[width=0.236\textwidth]{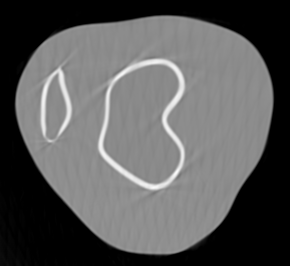}&
\includegraphics[width=0.236\textwidth]{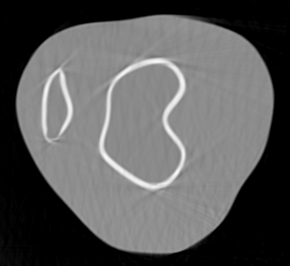}\\

\includegraphics[width=0.236\textwidth]{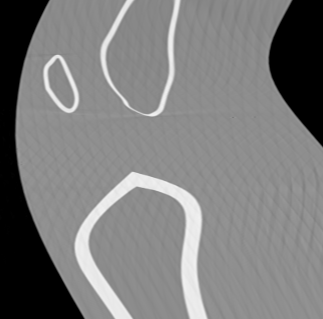}&
\includegraphics[width=0.236\textwidth]{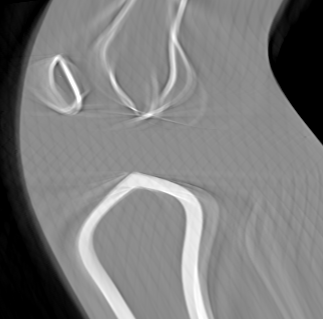}&
\includegraphics[width=0.236\textwidth]{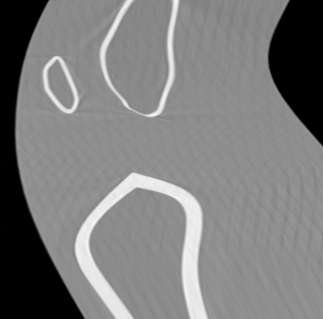}&
\includegraphics[width=0.236\textwidth]{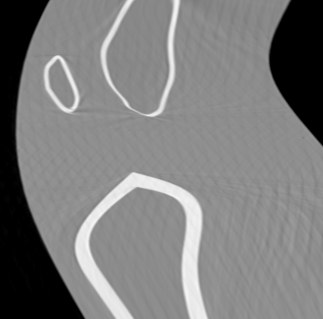}\\

(a) No motion&(b) Uncorrected&(c) Proposed&(d) Marker-based
\end{tabular}
    \caption{Exemplary slices of a reconstructed volume of a 30 degrees squat. First row: axial slice through shank, second row: axial slice through thigh, third row: sagittal slice. (a) scan without motion, (b) uncorrected case, (c) proposed method, (d) marker-based reference method. The motion artifacts clearly visible in the uncorrected case can be reduced by both the proposed and the reference method.} \label{fig-results}
\end{figure}

Exemplary slices of the resulting reconstructions are shown in Fig.~\ref{fig-results}.
A visual comparison of the results shows a similar reduction in streaking and blurring in the reconstructions of our proposed approach and the reference marker-based approach compared to the uncorrected case.
The average RMSE and SSIM values of the proposed approach and the marker-based reference approach compared to ground truth excluding background voxels are similar, see Table~\ref{tab1}.
Both methods resulted in a high average SSIM value of 0.99 (proposed) and 0.98 (reference) for 30 degrees squats and 0.98 (proposed) and 0.97 (reference) for 60 degrees squats.
The average RMSE of 0.02 for the proposed method was slightly lower than for the marker-based approach with on average 0.03.
Table~\ref{tab2} shows the average improvement in percent compared to the uncorrected case.
While the SSIM values show an improvement of 12-21\% for both proposed and reference approach, the RMSE improved on average by 68-70\% for the proposed method, and on average by 57\% for the marker-based approach.


\begin{table}[t]
    \caption{Average structural similarity (SSIM) index and root mean squared error (RMSE) over all subjects for 30$^\circ$ and 60$^\circ$ squats. Best values are printed bold.}
    \label{tab1}
    \centering

\setlength{\tabcolsep}{0pt} 

\begin{tabular*}{\textwidth}{@{\extracolsep{\fill}}lccccc}
\toprule
& \multicolumn{2}{c}{SSIM}
& \multicolumn{2}{c}{RMSE} &\\
\cmidrule{2-3}\cmidrule{4-6}
& 30$^\circ$ & 60$^\circ$ & 30$^\circ$ & 60$^\circ$ &\\
\midrule
uncorrected  & 0.881 $\pm$ 0.053    & 0.821 $\pm$ 0.098   & 0.070 $\pm$ 0.015   &   0.081 $\pm$ 0.020  &\\
proposed     & \textbf{0.989 $\pm$ 0.005}    & \textbf{0.984 $\pm$ 0.013}     & \textbf{0.022 $\pm$ 0.004}     & \textbf{0.024 $\pm$ 0.009}  &\\
marker-based &  0.980 $\pm$ 0.008       & 0.969 $\pm$ 0.017    & 0.029 $\pm$ 0.004   & 0.034 $\pm$ 0.008 &\\
\bottomrule
\end{tabular*}


    \caption{Average improvement in percent of structural similarity (SSIM) index and root mean squared error (RMSE) compared to the uncorrected case over all subjects for 30$^\circ$ and 60$^\circ$ squats. Best values are printed bold.}
    \label{tab2}
    \centering

\setlength{\tabcolsep}{0pt} 

\begin{tabular*}{\textwidth}{@{\extracolsep{\fill}}lccccc}
\toprule
& \multicolumn{2}{c}{SSIM}
& \multicolumn{2}{c}{RMSE} &\\
\cmidrule{2-3}\cmidrule{4-6}
& 30$^\circ$ & 60$^\circ$ & 30$^\circ$ & 60$^\circ$ &\\
\midrule
proposed     & \textbf{12.6 $\pm$ 6.7}    & \textbf{21.2 $\pm$ 14.3}     & \textbf{67.6 $\pm$ 7.1}     & \textbf{70.4 $\pm$ 5.5}  &\\
marker-based & 11.6 $\pm$ 6.3       & 19.3 $\pm$ 14.0    & 56.9 $\pm$ 7.0   & 57.1 $\pm$ 7.4 &\\
\bottomrule
\end{tabular*}
\end{table}

\section{Discussion and Conclusion}
The results presented in Fig.~\ref{fig-results} and Table~\ref{tab1} and \ref{tab2} show that the proposed method is able to estimate and correct for involuntary subject motion during a standing acquisition.
Compared to the case without motion, some artifacts are still visible and some double edges could not be restored.
A reason for this is that the motion applied for projection generation is a non-rigid motion, where the upper and lower leg can move individually.
The inertial sensor placed on the shank, however, is only able to estimate a rigid motion consisting of a 3D rotation and translation.
Therefore it is not possible to entirely restore image quality with one sensor, even though a clear improvement compared to the uncorrected case is observable.
To overcome this limitation, a second sensor could be placed on the thigh and both measurements could be combined to account for non-rigid motion. 

The reference approach tracking small metallic markers in the projection images for motion compensation also estimates a 3D rigid motion, thereby allowing for a fair comparison to our proposed approach.
The presented evaluation even shows slightly better results for the proposed approach in a qualitative as well as in a quantitative comparison.
Compared to the marker-based approach, where a sufficient number of markers has to be carefully placed around the knee, for our approach, only one or two sensors per leg are necessary to track the subject's motion.
This can help to facilitate and speed up the clinical process of weight-bearing imaging.

In this initial study, an optimal sensor in a well-controlled setting is assumed, however, in a real setting, some challenges will arise.
One potential issue will be noise in the sensor signals.
Furthermore, the gravitational signal is considerably larger than the motion to be estimated, which could pose a problem for an accurate estimation.
To evaluate the influence of measurement errors on the proposed method, in a subsequent study, errors like noise or gyroscope bias will be included in the simulation.

For this study, we assumed that the initial pose and velocity of the sensor at the beginning of the C-arm scan are known, while in a real setting, these values have to be estimated.
An approach for estimating the initial pose from the X-ray projection images was presented in \cite{Thies2019}.
If the sensor is placed close enough to the knee joint, it is also visible in the projection images without affecting image quality in the area of interest, and its metal components can be tracked.
Using the system geometry, an average position over the scan can be obtained, which then can be refined by the first projection images.
The initial velocity of the sensor could be estimated by assuming a zero mean velocity over the whole scan, since it has been shown that standing persons sway around their center of mass \cite{Abrahamova2008}.

We presented a novel approach for motion compensation in weight-bearing C-arm CT using inertial measurements.
This initial simulation is a first step towards a purely IMU-based compensation of motion in C-arm CT.
We showed the feasibility of our approach which was able to improve image quality and achieve results similar to a state-of-the-art marker-based motion compensation.
\subsubsection{Acknowledgment} Bjoern Eskofier gratefully acknowledges the support of the German Research Foundation (DFG) within the framework of the Heisenberg professorship programmme (grant number ES 434/8-1). The authors acknowledge funding support from NIH~5R01AR065248-03 and NIH Shared Instrument Grant No. S10 RR026714 supporting the zeego@StanfordLab. 
%
%
%
\bibliographystyle{splncs04}

\end{document}